\documentclass[11pt]{article}

\usepackage[utf8]{inputenc}
\usepackage{graphicx}
\usepackage{soul, color} 
\usepackage{mathtools}
\usepackage{microtype} 
\usepackage{booktabs}  
\usepackage{url}  
\usepackage{amsmath, amssymb, amscd, amsthm, amsfonts}

\usepackage[finalworkshop]{automl}
\usepackage{natbib}
\title{HPO: We won't get fooled again}

\author[1,2]{\nameemail{Kalifou Ren\'e Traor\'e}{kalifou@dlr.de}}

\author[2]{\nameemail{Andr\'es Camero}{andres.camerounzueta@dlr.de}}
\author[1,2]{\nameemail{Xiao Xiang Zhu}{xiaoxiang.zhu@dlr.de}}

\affil[1]{Data Science in Earth Observation, Technical University of Munich, Germany.}
\affil[2]{Remote Sensing Technology Institute (IMF), German Aerospace Center (DLR), Germany.}

\hypersetup{%
  pdfauthor={}, 
  pdftitle={},
  pdfsubject={},
  pdfkeywords={}
}

\begin{document}

\maketitle

\begin{abstract}

Hyperparameter optimization (HPO) is a well-studied research field.
However, the effects and interactions of the components in an HPO pipeline are not yet well investigated.
Then, we ask ourselves: \emph{Can the landscape of HPO be biased by the pipeline used to evaluate individual configurations?}
To address this question, we proposed to analyze the effect of the HPO pipeline on HPO problems using fitness landscape analysis.
Particularly, we studied the DS-2019 HPO benchmark data set, looking for patterns that could indicate evaluation pipeline malfunction, and relate them to HPO performance.
Our main findings are: 
(i) In most instances, large groups of diverse hyperparameters (i.e., multiple configurations) yield the same \emph{ill} performance, most likely associated with majority class prediction models; 
(ii) in these cases, a worsened correlation between the observed fitness and average fitness in the neighborhood is observed, 
potentially making harder the deployment of local-search based HPO strategies. 
Finally, we concluded that the HPO pipeline definition might negatively affect the HPO landscape.


\end{abstract}

\section{Introduction and Related Work}

Modern data-driven approaches dealing with large-scale data require domain, data science and technical expertise. 
The variety of application tasks (e.g., classification and object detection) often require designing models that are not necessarily 
reusable in other tasks, and this process is both resource-demanding and error-prone (\cite{Ojha2017}, \cite{Elsken2019}, \cite{Ren2021}). 
Thus, automating the design of ML pipelines, a.k.a. \emph{AutoML} (\cite{hutter2019automated}), is much desirable. 

AutoML is usually split into four main activities: Data preparation, feature engineering, model generation, and model estimation (\cite{he2021automl}). \emph{Hyperparameter optimization} (HPO, \cite{Bischl2021hpo}) is an important task in model generation. HPO
aims at automatically tuning the hyperparameters of learning algorithms, and as with all optimization problems, 
it is facing the process of minimizing/maximizing a target function (e.g., performance metric of the model) subject to a set of constraints. 
HPO is a well-studied field (\cite{Bischl2021hpo}), but the effects and interaction between 
the components of its pipeline is not yet well investigated. 
Recently, \cite{Pimenta2020autoMLfla} proposed to characterize the search space of AutoML pipelines using \emph{fitness landscape analysis} (FLA, \cite{pitzer2012comprehensiveFLA}). In the same line, \cite{traore2021footprint} proposed a FLA-base framework to characterize NAS problems, and applied it to a multi-sensor data fusion problem (\cite{traor2022landscapefusion}). Despite the great results and insights provided by these studies, the relation between HPO and  the rest of the HPO pipeline remains barely explored.

Therefore, in this study, we pose the following research question: 
\emph{\textbf{Can the landscape of HPO be biased by the pipeline used to evaluate individual configurations?}}
To address this question, we propose to study HPO in the context of AutoML using FLA. 
Particularly, using \emph{fitness distance correlation} (FDC, \cite{FDC_Metric_Jones}) and \emph{locality} (\cite{CLERGUE2018449}),  we aim at  patterns that arise from evaluation pipelines issues, and assess how they could alter the landscapes of HPO problems.
The results on the DS-2019 HPO benchmark (\cite{SharmaDS2019}) show the existence of large groups of diverse HP configurations that yield the same \emph{ill} fitness value. This \emph{illness} could be explained by the fitness metric selection (e.g., predictive accuracy), that \emph{induces} the generation of majority class predictors as a \emph{local optima configuration}.
A complementary analysis of locality shows that the resulting landscapes are more rugged, with lesser correlation between the observed fitness and the fitness in the neighborhood. In other words, these problems are hard to tackle using a local-search strategy.

The rest of the paper is as follows: The next section introduces the methodology used in the study, 
Section~\ref{section:results} presents results of landscape analysis on HPO problems, 
Section~\ref{section:conclusion} provides conclusions, and Section~\ref{section:limitations-impact} discusses the limitations and impact of this work.



\section{Methodology}\label{section:method}


Given a HPO problem, let $S$ be the HP configuration space,
$f$~the fitness function that assigns a value $f(x) \in \mathbb{R}$ to all configurations $x \in S$, and $N(x)$ a neighborhood operator that provides a structure to $S$. 
Then, the fitness landscape is defined as $\mathcal{L} = (S, f, N)$.

We are interested in exploiting the landscape definition to study the relation between the HPO landscape and the HPO pipeline, and check whether the pipeline may bias the HPO landscape. Particularly, we propose to use the FDC and \emph{locality} to characterize this relation. The motivation is that issues related to the evaluation pipeline should affect the fitness of configurations irrespectively of their configuration, and thus their distance to the optimum. In other words, repetitive or grouping patterns (such as lines) might appear when visualizing distributions of distances to the optimum. Moreover, the locality of the configuration space should be arbitrarily affected, i.e., some configurations should present an unexpected or \emph{random} behavior (in relation to the neighborhood).

Without loss of generality, we consider the problem of tuning the HPs of a fixed neural network architecture to perform a task (e.g., classification). Typically, the HP configuration space consists of mixed type features (continuous, discrete or categorical).
Thus, we propose to evaluate the distance between individuals using a dedicated similarity function, $\delta(x, y)$, introduced by \cite{Gower1971}. 
Then, we define a neighborhood function $N(x) = \{ y \in S \mid \delta(x, y) < \Delta \}$.

The FDC is often interpreted as a measure of the existence of search trajectories from randomly picked configurations to the known global optimum.
In practice, the FDC is not collected as a correlation score, but visualized as the distribution of fitness versus distance to the global optimum. It writes as:~$\operatorname{FDC}(f,x^*, S)=\{(\delta(x^*,y), f(y)) \mid \forall y \in S\}$, 
where $x^* \in S$ is the global optimum. 
On the other hand, \emph{locality} corresponds to the relationship between the observed fitness and the distribution of average fitness in the neighborhood (\cite{CLERGUE2018449}).

\section{Results}\label{section:results}

To evaluate the proposed methodology, we propose to analyze the \textbf{DS-2019} HPO benchmark data set. 
DS-2019 consists of a tabular benchmark for the scenario of tuning the HPs of a (fixed) convolutional neural network (CNN), a ResNet-18, 
on ten instances of CV classification. 
For each instance, 15 hyperparameters should be optimized, including the batch size, number of epochs and momentum, among others.

\subsection{Fitness Distance Correlation (FDC)}\label{subsection:FDC}
\vspace{-0.2cm}

\begin{figure}[!h]
\centering
    \includegraphics[height=0.1525\textheight]{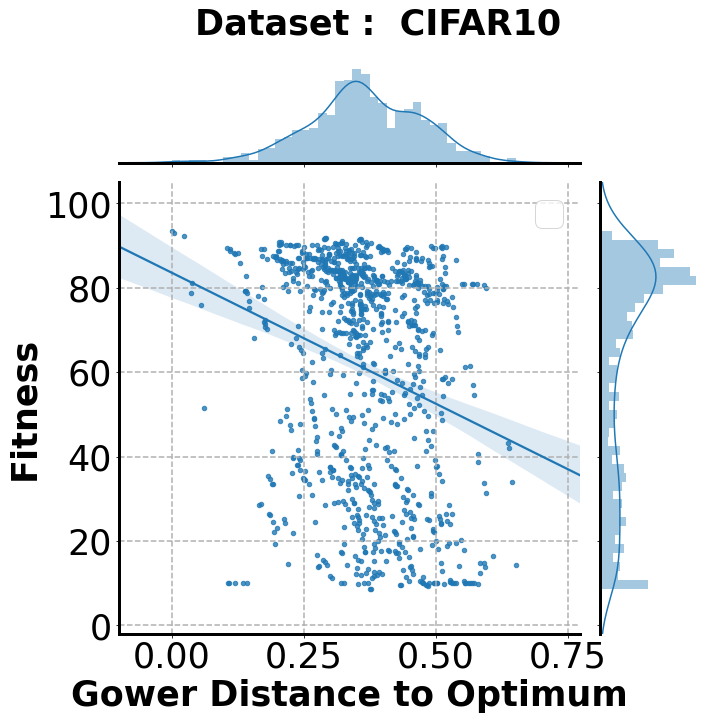}
    \includegraphics[height=0.1525\textheight]{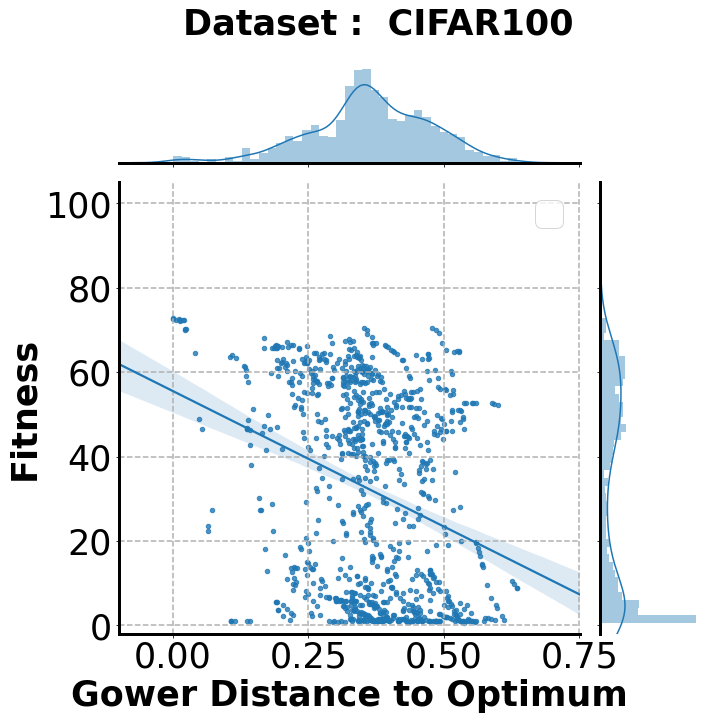}
    \includegraphics[height=0.1525\textheight]{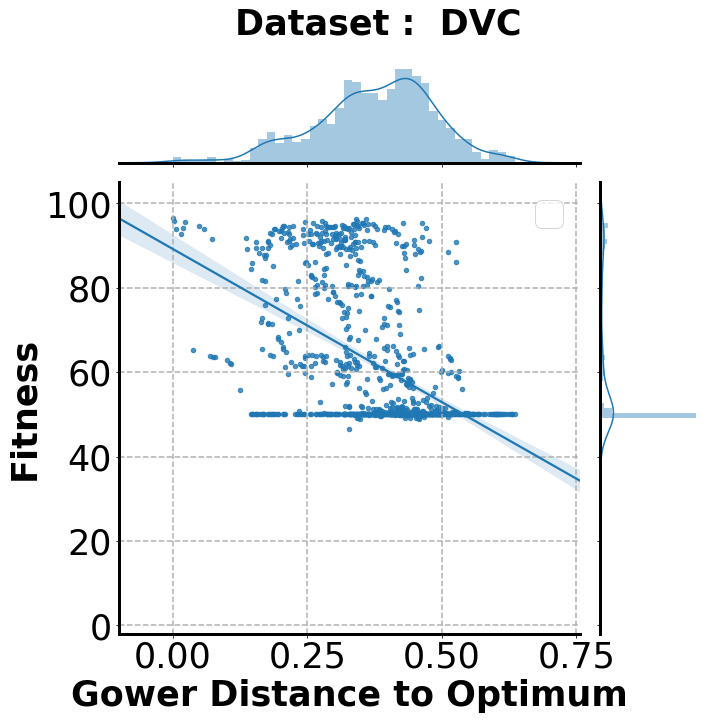}
    \includegraphics[height=0.1525\textheight]{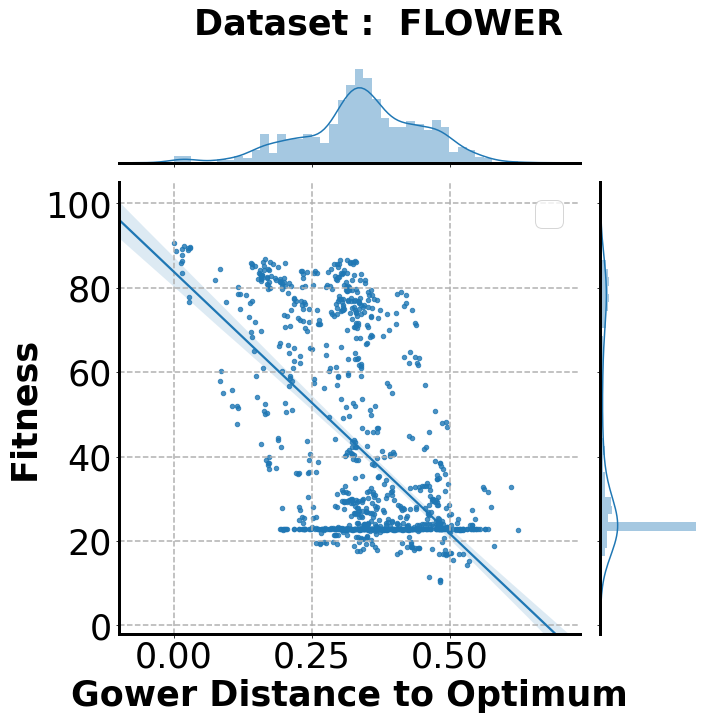}
    \includegraphics[height=0.1525\textheight]{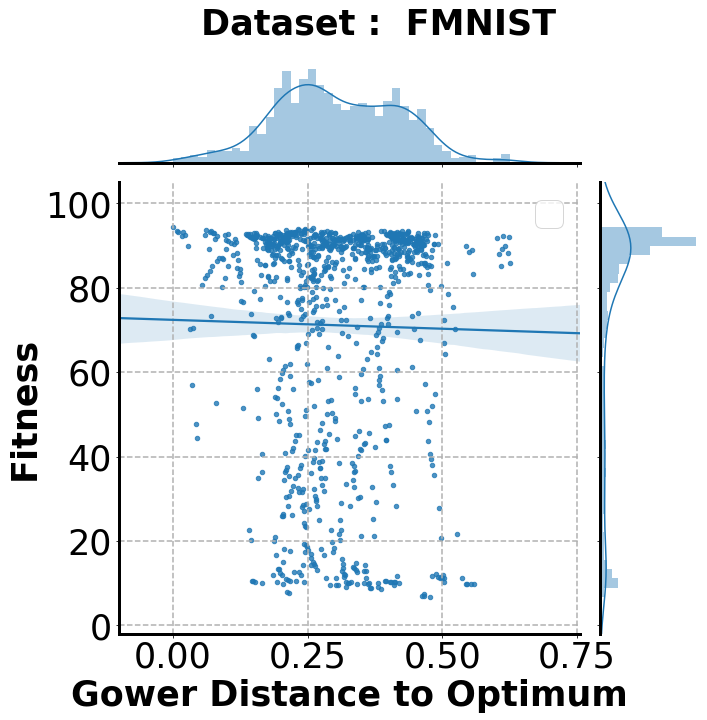}
    \includegraphics[height=0.1525\textheight]{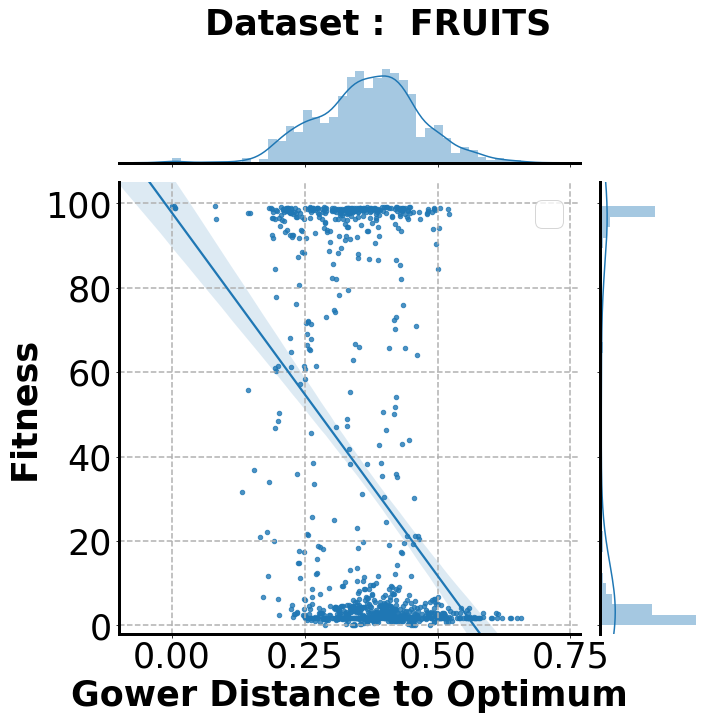}
    \includegraphics[height=0.1525\textheight]{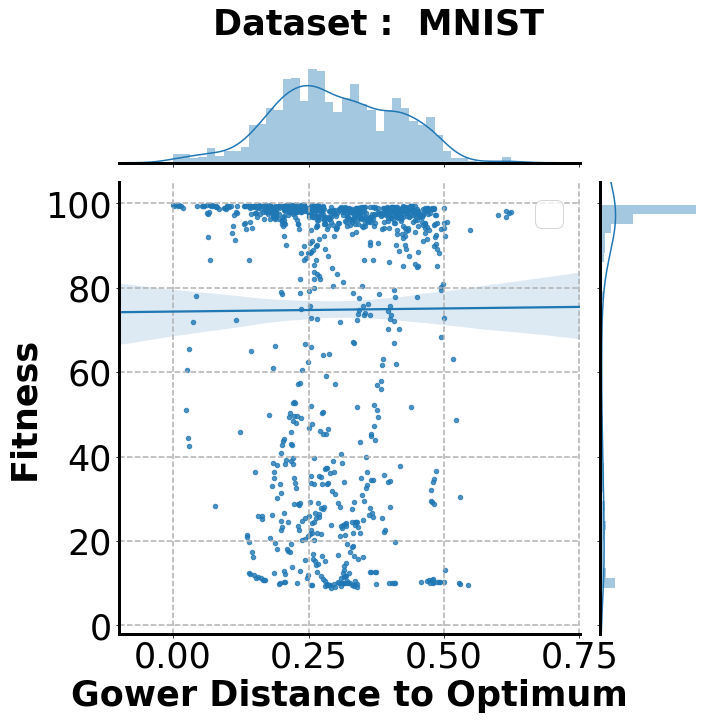}
    \includegraphics[height=0.1525\textheight]{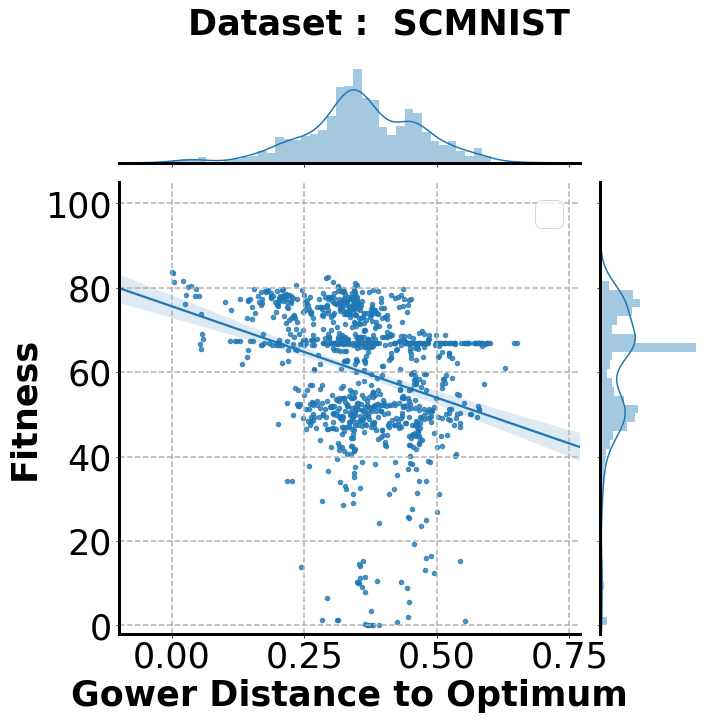}
    \includegraphics[height=0.1525\textheight]{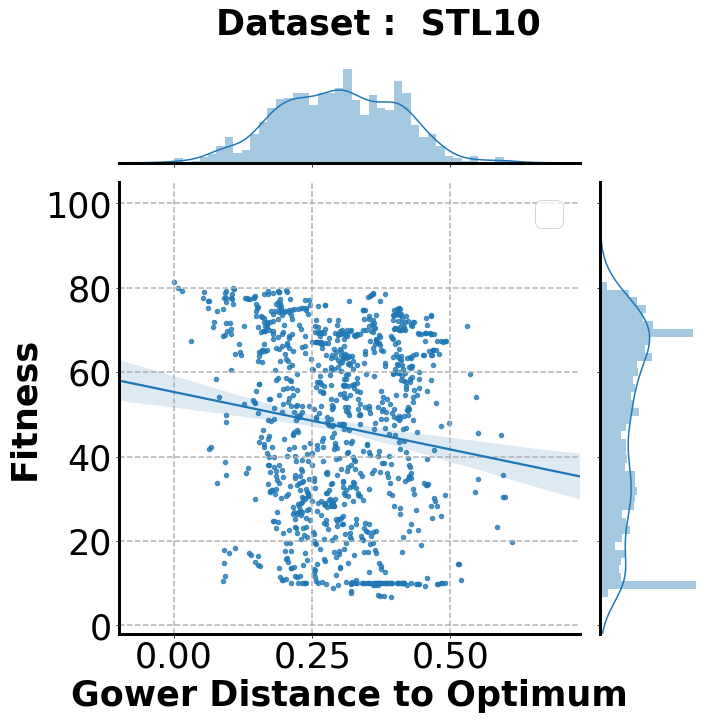}
    \includegraphics[height=0.1525\textheight]{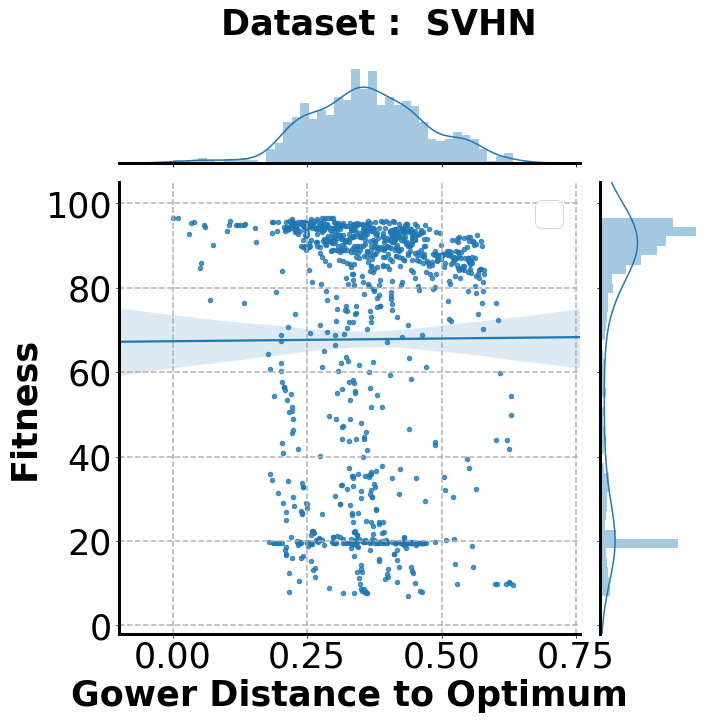}
    \caption{
    FDC plot for all of the instances, and the corresponding regression line in blue. 
    } 
    \label{fig:fdc}
\end{figure}

First, for each instance, we randomly sampled 1000 HP configurations, and computed the FDC (Figure~\ref{fig:fdc}).
Overall, the distances to the global optimum cover a wide range of values: the distribution of distances is wide and uniform for most instances. 
This suggests a large diversity in the HP configurations (with respect to the optimum), 
for the sample and potentially the whole configuration space.
Similarly, in most cases, the fitness also covers a wide range of values, i.e., all distributions appear to be multi-modal, 
with a principal mode for large fitness values (i.e., good configurations), and another mode for \emph{odd} values. 
We checked the data distribution for each instance, and we notice that the \emph{odd} modes could be correlated to the majority class. Note that the fitness metric used is the predictive accuracy. For example, on DVC it is around 50\%, FLOWER around 25\%, SCMNIST around 65\% and SVHN around 20\%. In particular, configurations are affected regardless of the distance to the optimum. In other words, very diverse configurations yield the same fitness value. This phenomenon could be attached to issues with the learning process, failing to properly fit the data and being stuck in poor local optima (i.e., majority class prediction), preventing them to reach the fitness that their HP configuration would normally yield.
Besides, there is no clear global correlation between the observed fitness and distance to the global optimum.
This could be caused by the multi-modal nature of the distributions of fitness. 


\vspace{-0.3cm}

\vspace{-0.6cm}

\subsection{Neighborhood}

\begin{figure}[h]
\centering
    \includegraphics[width=\textwidth]{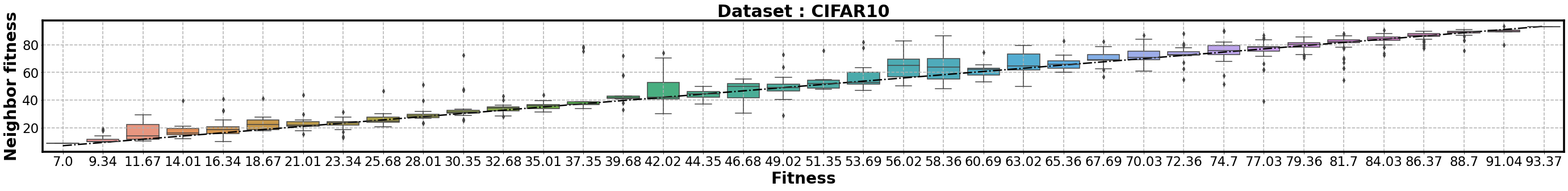}
    \includegraphics[width=\textwidth]{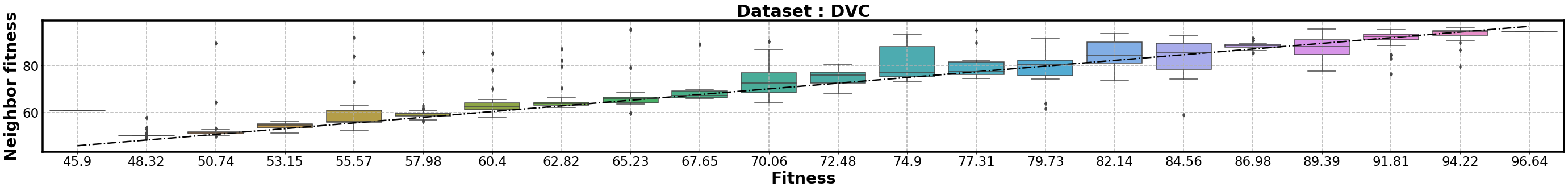}
    \includegraphics[width=\textwidth]{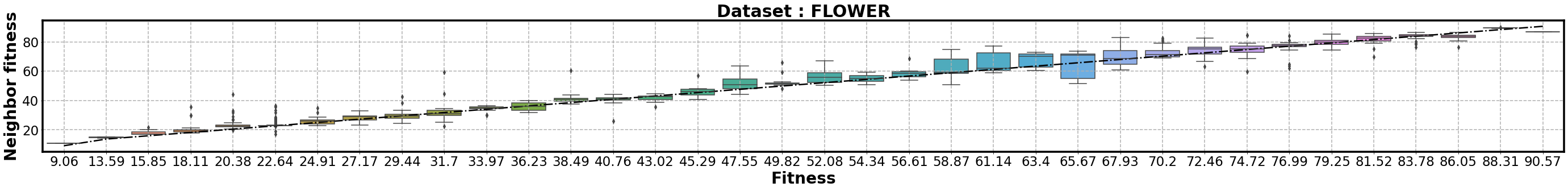}
    \includegraphics[width=\textwidth]{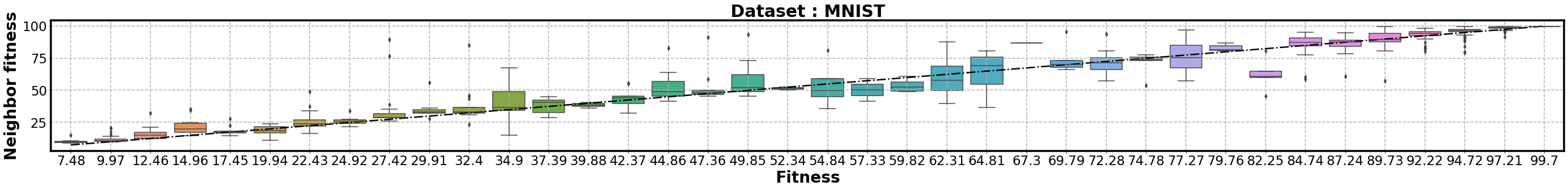}
    \includegraphics[width=\textwidth]{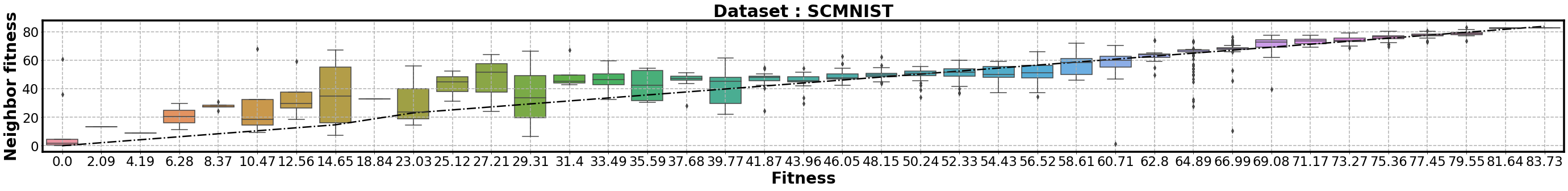}
    \includegraphics[width=\textwidth]{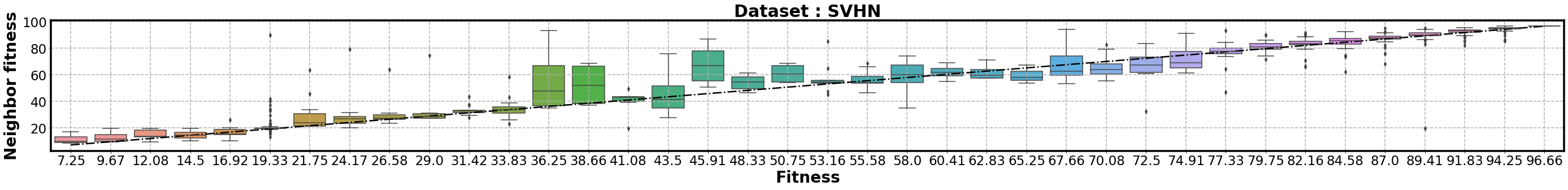}
    \caption{
    Distribution of the average fitness of neighbors as function of the observed fitness.
    } 
    \label{fig:neighborhood}
\end{figure}

Next, we seek to identify how the observed artifacts, i.e., the \emph{majority} class predictors, affect the locality of landscapes. 
Figure~\ref{fig:neighborhood} shows the distribution of average neighbor fitness as a function of the observed fitness, 
for six selected instances.  
The black dash-dotted line represents the bisector, 
i.e., the line connecting all points of equal value on both axis.
To generate the plots, we used the previously sampled configurations, and identified
the maximal pairwise distance (of any individual) to the optimum $\texttt{max}_{\text{dist}}$, 
and maximum observed fitness $\texttt{max}_{\text{fitness}}$. 
Given a constant $C=40$, we discretize the range of fitness values into intervals, where a step is equal 
to the maximum observed fitness $\texttt{max}_{\text{fitness}}$ divided by $C$. 
In order to decide if a configuration is a neighbor, we set $\Delta = \texttt{max}_{\text{dist}} / C$.    

Overall, we observe in most instances a strong correlation between the observed fitness and 
the average fitness in the neighborhood. Indeed, the box-plots are aligned with the bisector.
From the perspective of local search, 
it is easy to navigate the configuration space by consistently improving 
the fitness, from randomly distant and bad configurations, 
to configurations of high fitness.

Also, the instances with more uniform and wider distribution of fitness (Figure~\ref{fig:fdc}) tend to have a near perfect correlation.
On the other hand, the more the distributions are multi-modal and with peaky modes, 
the worse the correlation between the variables of interest.  
This suggests that the evaluation protocol could have an impact on the easiness and practicability of HPO landscapes, assessed by the correlation.


\vspace{-1.5cm}

\section{Conclusions and Future Work}\label{section:conclusion}\vspace{-0.3cm}
In this paper, we investigate if AutoML pipelines can negatively affect the landscape of HPO problems. 
More precisely, we address the following question: \emph{\textbf{Can the landscape of HPO be biased by the pipeline used to evaluate individual configurations?}}
To tackle this question, we have studied the fitness landscape of 10 HPO instances (from DS-2019 benchmark data set) using the \emph{fitness distance correlation} and \emph{locality}.

The FDC analysis shows unhealthy patterns in most HPO instances. For DVC, FLOWER, SCMNIST, and SVHN large groups of very diverse HP configurations with 
the same \emph{ill} fitness value are observed. These resulting peaks in fitness appear to be outliers in the respective distributions, potentially associated with majority class predictors. 
Looking at the locality (fitness versus fitness in the neighborhood), we observe two things: 
First, there is a correlation between both variables of interest, suggesting that an \emph{easy} path from randomly picked HP configurations could lead to the best performers, i.e., local-search may \emph{do the job}.
Second, for HPO problems negatively affected by the mentioned \emph{illness} (i.e., the majority class predictors),
the correlation between the current fitness and fitness in the neighborhood is worsened, indicating more rugged local landscapes.

Even though the \emph{majority class prediction} problem for models trained and evaluated using some metrics (e.g., accuracy) is well known, the results show that the problem may not be taken seriously into account. Thus, a great amount of resources is wasted when addressing HPO (i.e., many \emph{simple} majority class models are evaluated). Furthermore, the evidence show that \emph{\textbf{the landscape of HPO problems could be negatively affected by the evaluation pipeline being used}}.

Future work will further investigate the origin of such artifacts, 
as well as if they are present in other HPO problems (other scenarios or instances considering different fitness metrics).

\section{Limitations and Broader Impact Statement}\label{section:limitations-impact}

%

%

\subsection{Limitations: Additional Aspects of FLA for HPO}

The current FLA-based study on the effects and interactions of HPO in an AutoML pipeline lacks from the \emph{full picture}. Particularly, FDC and \emph{locality} provides a limited approximation to the landscape, thus additional aspects of FLA should be considered to further unveil these effects and interactions.
The FLA literature defines several tools to characterize different \emph{properties} of a landscape, including ruggedness, local optima cardinality, neutrality degree, and evolvability, among others~(\cite{pitzer2012comprehensiveFLA}).

\begin{figure}[!h]
\centering
    \includegraphics[width=\textwidth]{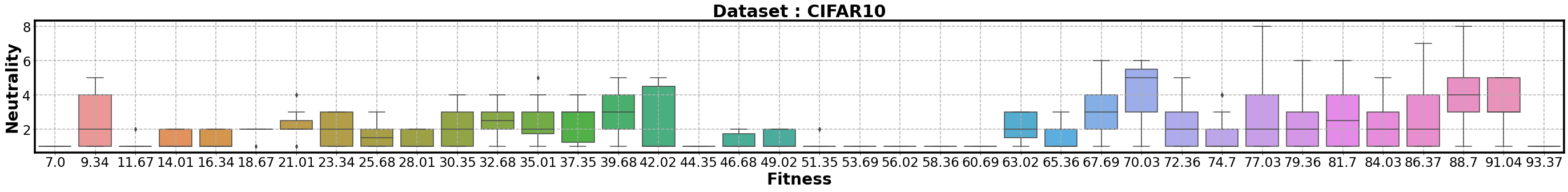}
    \includegraphics[width=\textwidth]{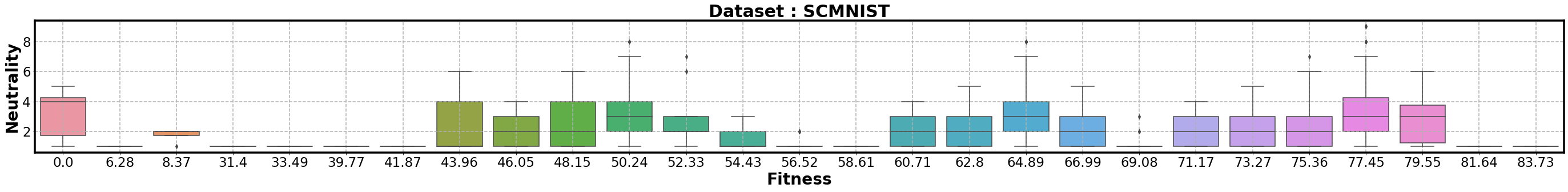}
    \caption{Neutrality degree as function of the observed fitness, for CIFAR-10 and SCMNIST.} 
    \label{fig:neutrality}
\end{figure}

For example, as a sneak peek of our future work, Figure~\ref{fig:neutrality} shows preliminary results of the neutrality degree (\cite{CLERGUE2018449}) as a function of the observed fitness for two instances. The~\emph{neutrality degree} is defined as $N_{d}(x) = \vert \{x^{'} \in N(x) \mid  \mid f(x^{'}) - f(x)\mid < \epsilon \} \vert$, and it is interpreted as the number of neighbors of $x$ that have a similar fitness. In this case, we set $\epsilon = \texttt{max}_{\text{fitness}} / C$.

Overall, the neutrality degree is equal or greater than one for most ranges of fitness values.
In order words, most configurations have at least one neutral neighbor. 
Also, note that the FDC and \emph{locality} results for CIFAR-10 are \emph{good}, while for SCMNIST, with a multi-modal distribution of fitness (Figure~\ref{fig:fdc}), coupled with lower \emph{local} correlation (i.e., between the fitness and the fitness in the neighborhood, Figure~\ref{fig:neighborhood}), the results are \emph{bad}.
Regarding CIFAR-10, the neutrality degree is on average consistently greater than two. In other words, most configurations have two or (many) more \emph{neutral} neighbors.
On the other hand, for SCMNIST, the neutrality degree is inconsistent and with lower values on average. In particular, $N_{d}$ is lower for fitness values ranging from $6.28$ to $43.98\%$, i.e., generally bad configurations have fewer neutral neighbors than \emph{mid} and \emph{good} configurations. Also, as expected, there is a huge number of neutral neighbors around the majority class prediction fitness ($65\%$)
As a summary, the evaluation pipeline \emph{malfunction} is responsible for an \emph{imbalance} landscape, i.e., the AutoML pipeline generates arbitrary \emph{peaks} of fitness (low $N_{d}$) in areas of expected continuous fitness.

\subsection{Landscape Analysis for HPO Benchmarks: A Tool for Troubleshooting?}

Generally speaking, the usefulness of FLA to characterize optimization problems is undoubtable. However, \emph{can we use FLA as a general tool for HPO troubleshooting?}. Furthermore, are the \emph{problems} observed in DS-2019 also present in other HPO benchmark data sets, e.g., KDD-2018 (\cite{Rijn2018Hyperparameter}) and YAHPO (\cite{pfisterer2021yahpo})? And are these \emph{artifacts} real \emph{problems}? One may argue that this is the behavior of the learning algorithm applied to a model and data set given a set of hyperparameters. Therefore, the achieved performance is just the \emph{expected} one. However, do we want to spend a non-neglectable amount of resources training and evaluating \emph{useless} or \emph{trivial} models? In the case of HPO benchmarks, it may be a good idea to have this behavior, as we may learn to avoid it on real applications. But, in our opinion, this issue should be avoided in real applications, in order to save time/resources. Thus, the insights provided by FLA should be taken into account to improve the design of HPO problems.



\bibliography{references} 



\newpage
\begin{acknowledgements}
Authors acknowledge support by the European Research Council (ERC) under the European Union's Horizon 2020 research and innovation program (grant agreement No. [ERC-2016-StG-714087], Acronym: \textit{So2Sat}), by the Helmholtz Association
through the Framework of Helmholtz AI [grant  number:  ZT-I-PF-5-01] - Local Unit ``Munich Unit @Aeronautics, Space and Transport (MASTr)'' and Helmholtz Excellent Professorship ``Data Science in Earth Observation - Big Data Fusion for Urban Research'' (W2-W3-100),  by the German Federal Ministry of Education and Research (BMBF) in the framework of the international future AI lab "AI4EO -- Artificial Intelligence for Earth Observation: Reasoning, Uncertainties, Ethics and Beyond" (Grant number: 01DD20001) and the grant DeToL. Authors also acknowledge support by DAAD for a Doctoral Research Fellowship.

Besides, we also would like to thank Lennart Schneider and Florian Pfisterer
for their valuable input and feedback, helping us improve the current paper.
\end{acknowledgements}

\section*{Reproducibility Checklist}

\begin{enumerate}
\item For all authors\dots
  \begin{enumerate}
  \item Do the main claims made in the abstract and introduction accurately
    reflect the paper's contributions and scope?
    \answerYes{Yes the claims in the abstract and introduction reflect the contributions in the paper.}
  \item Did you describe the limitations of your work?
    \answerYes{Yes, we did describe the limitations of the study.}
  \item Did you discuss any potential negative societal impacts of your work?
   \answerNA{Since our paper only provides an empirical study on the negatives impact of AutoML evaluations pipeline on algorithm deployment, 
   we believe that so far the societal impact is very limited.}  
  \item Have you read the ethics author's and review guidelines and ensured that your paper
    conforms to them? \url{https://automl.cc/ethics-accessibility/}
    \answerYes{Yes, we have read the guidelines and our paper conforms to them.}
  \end{enumerate}
\item If you are including theoretical results\dots
  \begin{enumerate}
  \item Did you state the full set of assumptions of all theoretical results?
    \answerNA{Our paper provides only an empirical study with no theoretical results. }
  \item Did you include complete proofs of all theoretical results?
     \answerNA{Our paper provides only an empirical study with no theoretical results. }
  \end{enumerate}
\item If you ran experiments\dots
  \begin{enumerate}
  \item Did you include the code, data, and instructions needed to reproduce the
    main experimental results, including all requirements (e.g.,
    \texttt{requirements.txt} with explicit version), an instructive
    \texttt{README} with installation, and execution commands (either in the
    supplemental material or as a \textsc{url})?
    \answerYes{Yes, the code, data and instructions are available.}
  \item Did you include the raw results of running the given instructions on the
    given code and data?
   \answerYes{Yes, raw results are available in a reproducible jupyter notebook.}
  \item Did you include scripts and commands that can be used to generate the
    figures and tables in your paper based on the raw results of the code, data,
    and instructions given?
    \answerYes{Yes, all are available in a reproducible jupyter notebook.}
  \item Did you ensure sufficient code quality such that your code can be safely
    executed and the code is properly documented?
    \answerYes{Yes, the code is clean and documented.}
  \item Did you specify all the training details (e.g., data splits,
    pre-processing, search spaces, fixed hyperparameter settings, and how they
    were chosen)?
    \answerNA{All hyperparameters are mentioned in the paper and appear in the code. 
    As we only do an empirical study of already trained models, the code mainly consists of a simple algorithms and plotting functions.}
  \item Did you ensure that you compared different methods (including your own)
    exactly on the same benchmarks, including the same datasets, search space,
    code for training and hyperparameters for that code?
    \answerYes{}{The algorithms and plotting functions were deployed on various data using the same setting.}
  \item Did you run ablation studies to assess the impact of different
    components of your approach?
    \answerNA{This does not apply to our simple empirical study.}
  \item Did you use the same evaluation protocol for the methods being compared?
    \answerNA{This does not apply to our simple empirical study.}
  \item Did you compare performance over time?
    \answerNA{This does not apply to our simple empirical study.}
  \item Did you perform multiple runs of your experiments and report random seeds?
    \answerNA{This does not apply to our simple empirical study.}
  \item Did you report error bars (e.g., with respect to the random seed after
    running experiments multiple times)?
    \answerNA{This does not apply to our simple empirical study.}
  \item Did you use tabular or surrogate benchmarks for in-depth evaluations?
    \answerYes{Yes, our study is based on analysing an tabular HPO benchmark suite.}
  \item Did you include the total amount of compute and the type of resources
    used (e.g., type of \textsc{gpu}s, internal cluster, or cloud provider)?
    \answerNA{This does not apply as we only analyze an existing HPO benchmark suite, and use methods and algorithms that are costless (CPU).}
  \item Did you report how you tuned hyperparameters, and what time and
    resources this required (if they were not automatically tuned by your AutoML
    method, e.g. in a \textsc{nas} approach; and also hyperparameters of your
    own method)?
    \answerNA{NA.}
  \end{enumerate}
\item If you are using existing assets (e.g., code, data, models) or
  curating/releasing new assets\dots
  \begin{enumerate}
  \item If your work uses existing assets, did you cite the creators?
    \answerYes{The creators of the DS-2019 HPO benchmark are cited.}
    %
  \item Did you mention the license of the assets?
     \answerNA{The benchmark suite is published as a paper at International Conference on Discovery Science, 
     and does not mention any license to be mentioned.} 
  \item Did you include any new assets either in the supplemental material or as
    a \textsc{url}?
    \answerNA{we do not use any additional material other than the previously mentioned.} 
  \item Did you discuss whether and how consent was obtained from people whose
    data you're using/curating?
    \answerNA{The benchmark suite is a tabular dataset for Hyperparameter performance on existing CV classification tasks. 
    In other words, the data used only collected algorithm performance, with no connection to any human activity.} 
  \item Did you discuss whether the data you are using/curating contains
    personally identifiable information or offensive content?
    \answerNA{This does not apply, as stated in the the previous answer.} 
  \end{enumerate}
\item If you used crowdsourcing or conducted research with human subjects\dots
  \begin{enumerate}
  \item Did you include the full text of instructions given to participants and
    screenshots, if applicable?
    %
    \answerNA{We did not crowdsource or conduct research with subjects.}
  \item Did you describe any potential participant risks, with links to
    Institutional Review Board (\textsc{irb}) approvals, if applicable?
     \answerNA{We did not crowdsource or conduct research with subjects.} 
  \item Did you include the estimated hourly wage paid to participants and the
    total amount spent on participant compensation?
    \answerNA{We did not crowdsource or conduct research with subjects.} 
  \end{enumerate}
\end{enumerate}

\appendix

\end{document}